\documentclass{article}

    \usepackage[preprint]{neurips_2024}
    \setcitestyle{numbers,square,comma}

\usepackage[utf8]{inputenc} 
\usepackage[T1]{fontenc}    
\usepackage{hyperref}       
\usepackage{url}            
\usepackage{booktabs}       
\usepackage{amsfonts}       
\usepackage{nicefrac}       
\usepackage{microtype}      
\usepackage{xcolor}         
\usepackage{graphicx}
\usepackage{float}
\usepackage{placeins}
\usepackage{caption}

\title{Enhancing Depression Diagnosis with Chain-of-Thought Prompting}

\author{%
  Elysia Shi \\
  Arcadia High School \\
  \texttt{contactelysias@gmail.com} \\
   \And
   Adithri Manda \\
   Monroe Township High School \\
   \texttt{adithri.manda16@gmail.com} \\
   \AND
   London Chowdhury \\
   Home-school \\
   \texttt{londonchowdhury@gmail.com} \\
   \And
   Runeema Arun \\
   North Creek High School \\
   \texttt{runeemaarun@hotmail.com} \\
   \AND
   Kevin Zhu \\
   Algoverse AI Research \\
   \texttt{kevin@algoverse.us} \\
   \And
   Michael Lam \\
   Algoverse AI Research \\
   \texttt{michael@algoverse.us} \\
}

\begin{document}

\maketitle

\begin{abstract}
 When using AI to detect signs of depressive disorder, AI models habitually draw preemptive conclusions. We theorize that using chain-of-thought (CoT) prompting to evaluate Patient Health Questionnaire-8 (PHQ-8) scores will improve the accuracy of the scores determined by AI models. In our findings, when the models reasoned with CoT, the estimated PHQ-8 scores were consistently closer on average to the accepted true scores reported by each participant compared to when not using CoT. Our goal is to expand upon AI models' understanding of the intricacies of human conversation, allowing them to more effectively assess a patient's feelings and tone, therefore being able to more accurately discern mental disorder symptoms; ultimately, we hope to augment AI models' abilities, so that they can be widely accessible and used in the medical field.
\end{abstract}

\section{Introduction}
Depression is a damaging mental disorder affecting hundreds of millions of people worldwide \cite{who2021depression}. It's characterized by constant feelings of sadness, lost interest in enjoyed activities, and degradation of an individual's person and can lead to self-harm, abuse, and in too many cases, suicide. Orthodox diagnostic methods begin with either standardized questionnaires that are often subjective and time-consuming or mental status examinations which are not accessible for many people due to various reasons, some being economic and insurance statuses \cite{batterham2016psychometric,plos2016fpa,compton2004diagnostic}.

The growing use of AI chatbots has made waves in mental health diagnostics, specifically concerning depressive disorders. Machine learning (ML) models have the potential to precisely distinguish subtle signals in human behavior. Large language models (LLMs) such as GPT-4 and BERT demonstrate the increased ability to dissect clinical data for use in bettering mental health diagnostics \cite{bmc_informatics}. These models can analyze great amounts of data including clinical notes, patient interviews, and social media posts, to determine patterns indicating depression. 
\subsection{Related works}
ML models, most notably LLMs, have assisted clinical psychologists in detecting symptoms of depression. In this manner, MIT Media Lab \cite{mitmedialab_depression} reports reveal that LLMs can accurately perceive depressive signs from social media posts. Moreover, LLMs have been used to analyze text from a variety of sources and reveal patterns that align with symptoms to indicate mental health issues \cite{johnson2024llms, liu2023llms}. Specifically, BERT was applied to electronic health records, achieving a precision of 90 percent in predicting depressive episodes of patients \cite{kim2021bert}.

Despite these advancements, several gaps remain. These types of models much of the time have issues regarding data privacy, biases, and ethical implications \cite{miller2024methodologies}. Additionally, a major struggle lies in the interpretability of ML models, including LLMs. These models often function as "black boxes", unable to understand and explain the reasoning behind their predictions. Their conclusions when diagnosing depression often arise with no indication of any human-understandable thought process, a feature characteristic of human diagnoses \cite{pmc10982476}. 

To address these issues, our research incorporates CoT prompting to enhance the accuracy and robustness of conclusions \cite{wei2022chain}. CoT prompting helps improve the model's internal decision-making process by simulating a step-by-step logical sequence, leading to better outcomes. By integrating CoT reasoning, we aim to improve the model’s decision-making ability.

\section{Methods}
\label{gen_inst}

 Our study uses the qualitative data of the DAIC-WOZ dataset \cite{gratch2014distress, devault2014simsensei}, to help train the model on depression symptoms. We go through the selection as well as pre-processing of the dataset, the training and developing of the models, and the incorporation of Cot prompting, and then we evaluate whether the model determined a PHQ-8 score closer to the actual score of the patient recorded in the data.

\subsection{Data collection and processing}

We used the dataset crafted by the University of Southern California called DAIC-WOZ. The DAIC-WOZ dataset is a collection of interviews with an animated virtual interviewer, Ellie, who was controlled by a human interviewer in a separate room. Ellie would ask patients a set of questions and transcribe the responses. The patient's facial expression data was also recorded. Each interviewee would subsequently fill out the PHQ-8; their PHQ-8 score, a valid diagnostic and severity measure for depressive disorders, would later be calculated according to the PHQ-8 score rubric \cite{kroenke2009phq8}. 

For this experiment, we assumed that each patient truthfully filled out the PHQ-8. The questionnaire has a set of questions, as shown in (Table \ref{tab:phq-8}). The patient will rate how well each individual phrase applied to their lives in the last 2 weeks on a scale of 0 to 3, with not at all being 0, several days being 1, more than half the days being 2, and nearly every day being 3. The number values for each question are then added up to get the total PHQ-8 score. The higher the score, the higher the severity of the interviewee's depression, if present at all. The data we utilized for the experiment was comprised of participants' PHQ-8 scores and interview transcripts.

\begin{table}[h!]
    \centering
    \begin{tabular}{|c|c|} \hline 
    \textbf{Question} & \textbf{Score} \\
    \hline
    1. Little interest or pleasure in activities & 0--3 \\
    \hline
    2. Feeling down or depressed & 0--3 \\
    \hline
    3. Trouble sleeping & 0--3 \\
    \hline
    4. Feeling tired or low energy & 0--3 \\
    \hline
    5. Poor appetite or overeating & 0--3 \\
    \hline
    6. Feeling bad about yourself & 0--3 \\
    \hline
    7. Trouble concentrating & 0--3 \\
    \hline
    8. Moving or speaking slowly or restlessness & 0--3 \\
    \hline
    \end{tabular}
    \vspace{12pt} 
    \caption{Sample PHQ-8 Rubric}
    \label{tab:phq-8}
\end{table}

This study followed an experimental study design. We utilized the same pre-trained model, OpenAI 3.5 turbo, for both the control and experimental test. In the control, the model was given participant interview transcript data and instructed to assign scores for each PHQ-8 category and calculate the total score. In the experimental test, the model was once again given transcript data and instructed to assign scores, only this time being prompted to use CoT reasoning.

\subsection{Procedure}
We created a control group (Assigner A), which we first fed the PHQ-8 rubric. A transcript was then passed to the model before prompting for a total PHQ-8 score as well as a score on each rubric question. The experimental group (Assigner B) was passed the same rubric and transcripts but with the addition of CoT prompting. We used zero-shot CoT prompting \cite{kojima2022large}. This method involves telling the model: “Let’s think step by step” in addition to the original prompt. 

\FloatBarrier 

\begin{figure}[H] 
    \centering
    \includegraphics[width=1\textwidth]{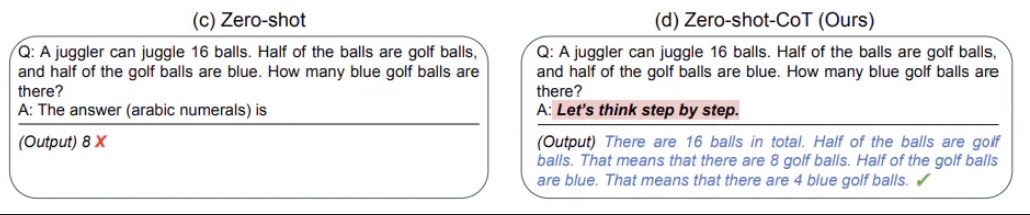} 
    \caption{Illustration of zero-shot chain of thought prompting \cite{kojima2022large} }
    \label{fig:zero_shot_prompting_word_example}
\end{figure}

\FloatBarrier 

\section{Results}
As seen in the graph, across all categories of behavior, Assigner B utilizing CoT reasoning generated scores far closer to the patient's accepted true PHQ-8 score as seen in lower average point differences 100 percent of the time.
\FloatBarrier 

\begin{figure}[H] 
    \centering
    \includegraphics[width=1\textwidth]{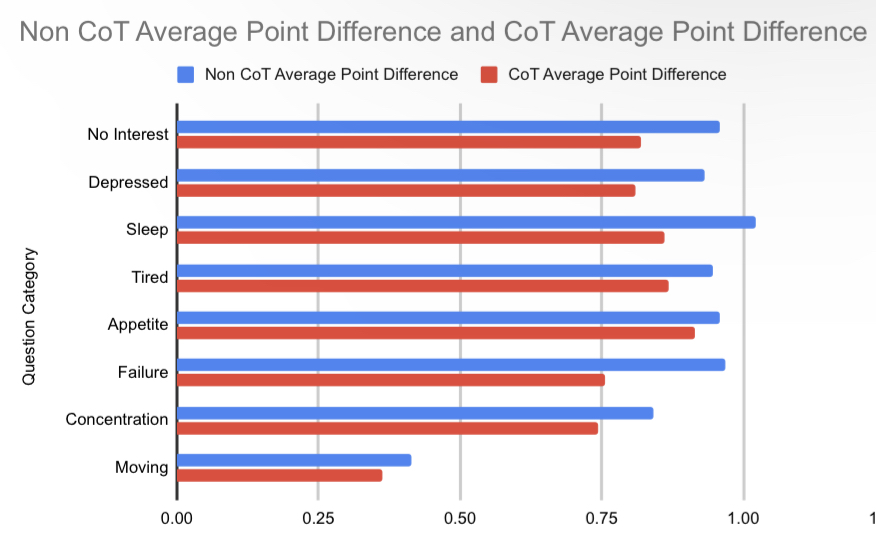} 
    \captionsetup{justification=centering}
    \caption{Average point difference of Assigner A scores and the true scores compared to the average point difference of Assigner B scores and the true scores}
    \label{fig:A_B_Assigner_Compare_Graph}
\end{figure}

\FloatBarrier 
\section{Discussion}
\subsection{Interpretation}
Our research aimed to determine and develop the accuracy of AI models in detecting symptoms of depression and assigning PHQ-8 scores with the use of chain of thought prompting and reasoning. The results demonstrate that implementing CoT prompting did in fact provide PHQ-8 scores closer to the actual score compared to the model without this prompting. After running the simplest of statistical tests, Assigner B's average point difference was not shown to be significantly lesser than that of A's; however, more in-depth statistical analysis will likely prove statistical significance.  Thus the aim of the model is supported: integrating CoT prompting into models does enhance their performance when surveying for depression indicators.

The improvement in accuracy is a product of the model’s ability to process and consider each aspect of the patient’s responses in a more structured and logical manner. This approach lines up with how human clinicians approach depression diagnosis: looking for different symptoms and breaking them down to acknowledge their impact on the overall disorder. This paper’s findings support the hypothesis that utilizing a logical reasoning process lets models better and more effectively analyze and interpret the qualitative data, specifically from patient interviews, which leads to better decision-making accuracy when determining signs of depression.

\subsection{Implications}
The findings in this paper have significant implications for mental health diagnostics and the potential use of AI in clinical practice. Utilizing CoT prompting in models improves the accuracy of PHQ-8 score predictions, making it a valuable tool to make mental health diagnoses, and potentially make mental health resources more accessible to the public.

Several gaps in this research highlight the need for further investigation to apply this model to potential diagnostic options. Since this study focused on the DAIC-WOZ dataset, which does not capture the diversity of patient responses, future studies should use larger and more varied datasets to ensure generalizability. Additionally, the questions in this dataset were insufficient for a robust evaluation. While CoT prompting enhances interpretation, its underlying mechanisms require further exploration to optimize chain-of-thought reasoning in different models.

Addressing these gaps is crucial for advancing the application of these findings, whether clinical or not, to ensure the model's optimal implementation. This experiment utilized a simpler GPT model; future research should explore CoT prompting with other models and advanced GPT versions. Testing variations like few-shot CoT could further optimize CoT usage. Integrating CoT prompting with advancements like explainable AI could enhance AI's effectiveness and acceptance in diagnosis. Policymakers and providers must develop guidelines and training programs to ensure the safe and effective use of AI diagnostic tools as AI's potential in healthcare grows.

\subsection{Limitations}
Currently, strong assumptions are being made regarding the accuracy of the interviews (in the DAIC-WOZ dataset) utilized as well as the dependence on data that is not burdened with errors, inaccuracies, and irrelevant information. In conjunction with this, there is also the assumption being made that the model has sufficient specifications, particularly the fact that the interviews used to train the model could have linguistic patterns that might vary when applied to different groups of people. There is also the assumption that CoT prompting will improve model performance, when the performance may even become worse with the complexity adding new sources of error and increasing the computational difficulty. Additionally, the use of AI in mental health diagnostics raises ethical concerns regarding data privacy, potential biases in data, and the implications of automated decision-making in patient care. By addressing and reflecting on our limitations, we aim to showcase transparency that will build credibility and trust in our AI models used for depression diagnosis.
\section{Conclusion}
Our research demonstrates that utilizing CoT prompting with AI models improves the accuracy of PHQ-8 score predictions for assessing symptoms of depression. By using logical reasoning processes, our model can more effectively analyze qualitative data, leading to more reliable and easier-to-interpret outcomes. This approach addresses the important challenges of AI use in mental health diagnostics, specifically transparency and ethical considerations, so that models that use CoT prompting can be utilized to make mental health care more accessible, in and out of clinical environments.

\bibliographystyle{abbrvnat}
\bibliography{refs} 

\end{document}